\pdfoutput=1

\documentclass[11pt]{article}

\usepackage[]{acl}

\usepackage{times}
\usepackage{latexsym}

\usepackage{tikz}
\usepackage{booktabs}       %
\usepackage{fdsymbol}
\usepackage{color}
\usepackage{colortbl}
\usepackage{multirow}
\usepackage{hyperref}

\usepackage[T1]{fontenc}

\usepackage[utf8]{inputenc}

\usepackage{microtype}

\usepackage{tabularx}
\usepackage{framed}
\definecolor{grey}{gray}{0.9}

\newcommand{\citeposs}[1]{\citeauthor{#1}'s (\citeyear{#1})}

\newcommand{\slashn}{\textbackslash n}

\newcommand{\Hrule}[3][.]{%
  \par\addvspace{#2}%
  \begingroup\color{#1}%
  \hrule
  \endgroup
  \addvspace{#3}%
}

\newcolumntype{G}{>{\columncolor{grey}}X}

\newenvironment{cframed}[1][gray!40]
  {\def\FrameCommand{\fboxsep=\FrameSep\fcolorbox{#1}{white}}%
    \MakeFramed {\advance\hsize-\width \FrameRestore}}
  {\endMakeFramed}

\title{\textsc{WinoPron}: Revisiting English Winogender Schemas \\ for Consistency, Coverage, and Grammatical Case}

\author{
  Vagrant Gautam\textsuperscript{1} \,
  Julius Steuer\textsuperscript{1} \,
  Eileen Bingert\textsuperscript{1} \\
  \textbf{Ray Johns\textsuperscript{2}} \,
  \textbf{Anne Lauscher\textsuperscript{3}} \,
  \textbf{Dietrich Klakow\textsuperscript{1}}
  \\
  \textsuperscript{1}Saarland University, Germany \,
  \textsuperscript{2}Independent Researcher, USA \\
  \textsuperscript{2}Data Science Group, University of Hamburg, Germany 
  \\
  \texttt{vgautam@lsv.uni-saarland.de}
  \\
}

\begin{document}
\maketitle

\begin{abstract}
While measuring bias and robustness in coreference resolution are important goals, such measurements are only as good as the tools we use to measure them.
Winogender Schemas~\citep{rudinger-etal-2018-gender} are an influential dataset proposed to evaluate gender bias in coreference resolution, but a closer look reveals issues with the data that compromise its use for reliable evaluation, including treating different pronominal forms as equivalent, violations of template constraints, and typographical errors.
We identify these issues and fix them, contributing a new dataset: \textsc{WinoPron}.
Using \textsc{WinoPron}, we evaluate two state-of-the-art supervised coreference resolution systems, SpanBERT, and five sizes of FLAN-T5, and demonstrate that accusative pronouns are harder to resolve for all models.
We also propose a new method to evaluate pronominal bias in coreference resolution that goes beyond the binary.
With this method, we also show that bias characteristics vary not just across pronoun sets (e.g., \textit{he} vs. \textit{she}), but also across surface forms of those sets (e.g., \textit{him} vs. \textit{his}).
\end{abstract}

\section{Introduction}
Third-person pronouns (\emph{he}, \emph{she}, \emph{they}, etc.) help us refer to people in conversation.
Since they mark referential gender in English, gender bias affects how coreference resolution systems map these pronouns to people.
\citet{rudinger-etal-2018-gender} demonstrated this by introducing Winogender Schemas, a challenge dataset to evaluate occupational gender bias in coreference resolution systems.
The dataset has become popular due to its careful construction;
it has been translated to other languages~\citep{hansson-etal-2021-swedish,stanovsky-etal-2019-evaluating} and used in framings beyond coreference resolution,  e.g., to evaluate natural language inferences~\citep{poliak-etal-2018-collecting} and intrinsic bias in language models~\citep{kurita-etal-2019-measuring}.\looseness=-1

\begin{figure}[t!]
    \centering
    \includegraphics[width=\linewidth]{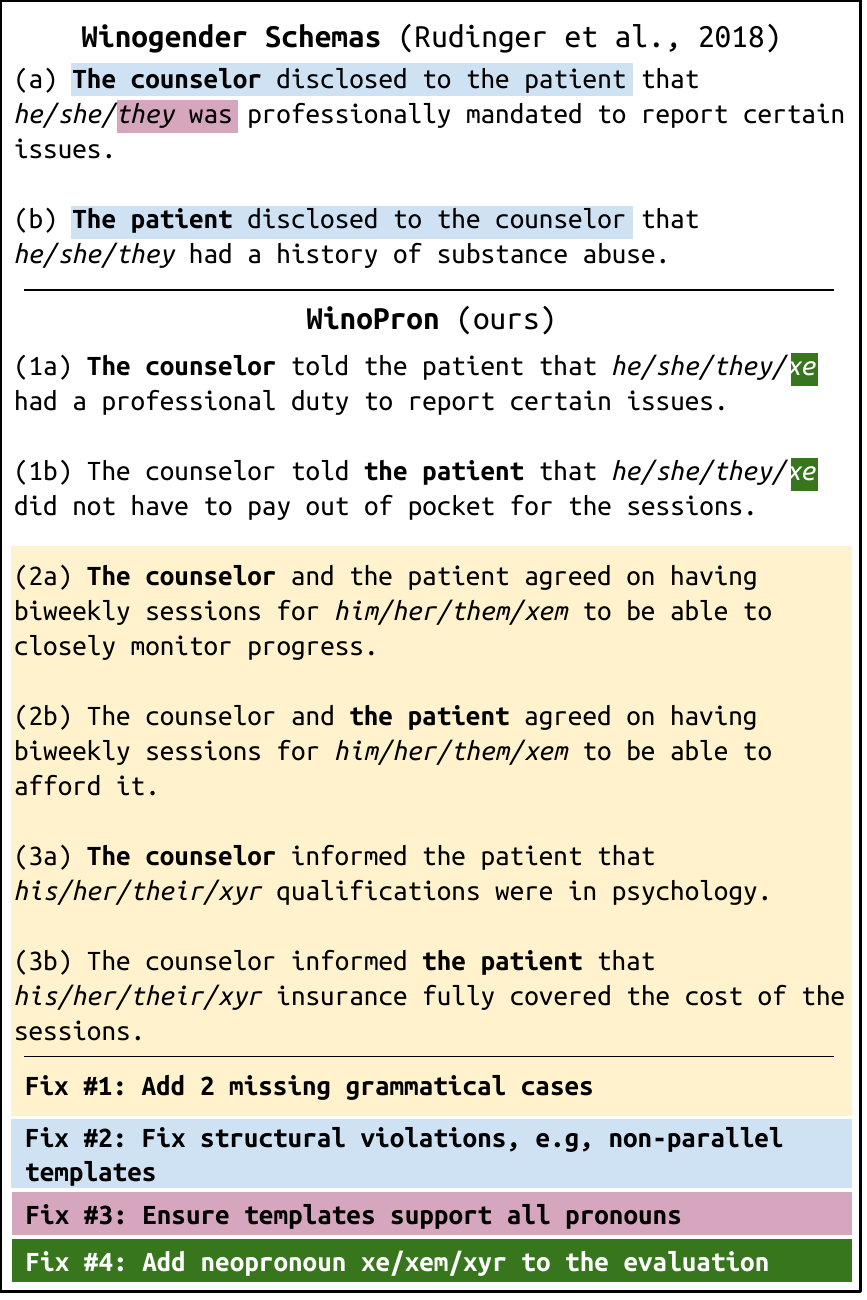}
    \caption{Problems with Winogender Schemas that we fix in our new coreference resolution dataset, \textsc{WinoPron}. Correct antecedents appear in \textbf{bold}.}
    \label{fig:one}
\end{figure}

However, a closer look at the dataset reveals weaknesses that compromise its use for reliable evaluation (see Figure \ref{fig:one}), which we hypothesize would affect both performance and bias evaluation.

In this paper, we identify issues with the original dataset and fix them to create a new dataset we call \textsc{WinoPron} (\S\ref{sec:winopron}).\footnote{Data and code available at \href{https://github.com/uds-lsv/winopron}{github.com/uds-lsv/winopron}.}
We then empirically show how our fixes affect coreference resolution system performance (\S\ref{sec:performance-consistency}) as well as bias (\S\ref{sec:bias}), with a novel method we propose to evaluate pronominal bias in coreference resolution that goes beyond the binary and focuses on linguistic rather than social gender~\citep{cao-daume-iii-2021-toward}.

\noindent Our fixes reveal that grammatical case, which we balance for in \textsc{WinoPron}, does indeed matter for both performance and bias results;
accusative pronouns are harder to resolve than nominative or possessive pronouns,
and system pronominal bias is not always consistent across different grammatical cases of the same pronoun set.
We find that singular \textit{they} and the neopronoun \textit{xe} are extremely hard for supervised coreference resolution systems to resolve, but surprisingly easy for FLAN-T5 models of a certain size.
We put forth hypotheses for these patterns and look forward to future work testing them.\looseness=-1

\section{Background: Winogender Schemas}
\label{sec:background}

\begin{figure}[t!]
    \begin{cframed}
    \noindent (a) The cashier told \textbf{the customer} that \textit{his / her / their} card was declined.
    
    \noindent (b) \textbf{The cashier} told the customer that \textit{his / her / their} shift ended soon.
    \end{cframed}
    \vspace{-3mm}
    \caption{Winogender Schemas for \textit{cashier}, \textit{customer} and possessive pronouns, with the antecedent bolded.}
    \label{fig:winogender}
    \vspace{-3mm}
\end{figure}

Winogender Schemas~\citep{rudinger-etal-2018-gender} are a widely-used dataset consisting of paired sentence templates in English, with slots for two human entities (an occupation and a participant), and a third person singular pronoun.
As Figure \ref{fig:winogender} shows, the second part of each template disambiguates which of the two entities the pronoun uniquely refers to, similar to Winograd schemas~\citep{levesque_et_all_2012}.
Changing the pronoun (e.g., from \textit{his} to \textit{her}) maintains the coreference, allowing us to measure whether coreference resolution systems are worse at resolving certain pronouns to certain entities.
\citet{rudinger-etal-2018-gender} use the gendered associations of these pronouns to show that gender bias affects coreference resolution performance.

The entities consist of 60 occupation-participant pairs (e.g., \textit{accountant} is paired with \textit{taxpayer}).
A pair of templates is created for each occupation-participant pair, resulting in a total of 120 unique templates.
The template pairs are designed to be parallel until the pronoun, such that only the ending can be used to disambiguate how to resolve the pronoun: it should resolve to the occupation in one template, and to the participant in the other.
Each template can be instantiated with three pronoun sets (\textit{he}, \textit{she}, and singular \textit{they}), for a total of 120 x 3 = 360 sentences for evaluation.\looseness=-1

\section{WinoPron Dataset}
\label{sec:winopron}

Although Winogender Schemas are established in the coreference resolution literature, we find issues with the dataset that compromise its use for reliable evaluation (see Figure \ref{fig:one} for examples).
We first motivate these issues and our fixes, and then describe how we create and systematically validate our new dataset, \textsc{WinoPron}.

We mostly reuse the occupation-participant pairs from Winogender Schemas (see Appendix \ref{sec:occupation-list} for the full list of pairings), but add 240 templates to cover missing grammatical cases, for a total of 360 templates.
We also include a neopronoun set (\textit{xe/xem/xyr}), giving us 360 templates x 4 pronoun sets = 1,440 sentences for evaluation.

\begin{table}[t!]
  \centering
  \begin{tabularx}{\linewidth}{Xrr}
    \toprule 
    \textbf{Grammatical case} &
    \textbf{WS} & \textbf{WP} \\
    \midrule
    Nominative (\textit{he}, \textit{she}, \textit{they}, \textit{xe}) & 89 & 120 \\
    Accusative (\textit{him}, \textit{her}, \textit{them}, \textit{xem}) & 4 & 120 \\
    Possessive (\textit{his}, \textit{her}, \textit{their}, \textit{xyr}) & 27 & 120 \\
    \bottomrule
  \end{tabularx}
   \caption{Number of templates per grammatical case in Winogender Schemas (WS) and \textsc{WinoPron} (WP).}
 \label{tab:balancing-case}
 \vspace{-3mm}
\end{table}

\subsection{Issues and Solutions}
\label{sec:issues-solutions}

\paragraph{Support for 3 Grammatical Cases}
We hypothesize that systems have different performance and bias characteristics with pronouns in different grammatical cases.\footnote{Here, we mean the surface form of the pronoun.} 
However, as Table \ref{tab:balancing-case} shows, Winogender Schemas have a variable number of pronouns per grammatical case, and treat them all as equivalent.
To enable more granular evaluation, we balance this distribution in \textsc{WinoPron}.

\paragraph{Consistency Fixes}
Winograd-like schemas have strict structural constraints so that models cannot inflate performance through heuristics.
However, when analyzing Winogender Schemas, we found constraint violations, e.g., non-parallel paired templates.
We fixed these along with typographical errors to ensure robust and reliable evaluation.

\paragraph{Support for All English Pronouns}
For a controlled evaluation comparing pronouns, it is common to use templates that only vary the pronoun.
However, 17\% of Winogender Schemas must be modified to work with singular \textit{they} due to its different verbal agreement (``he was'' but ``they were'').
To ensure a fair comparison between pronouns, we modify these templates to work with any pronouns.

\paragraph{Single-Entity Versions}
When evaluating large language models on coreference resolution when they have not explicitly been trained for it, poor performance could mean that the model simply cannot perform the task (with a given prompt).
In its current form, Winogender Schemas do not allow us to disentangle \textit{why} bad model performance is bad.
In \textsc{WinoPron}, we create single-entity sentences that are parallel to the traditional, more complex double-entity sentences, for a simple setting to test this, and a useful baseline for all systems.

\subsection{Data Creation}
\label{sec:data-creation}

Two authors with linguistic training iteratively created sentence templates until we reached consensus on their grammaticality and correct, unique coreferences.
We found template construction to be particularly challenging and time-consuming, due to ambiguity and verbal constraints.

\paragraph{Ambiguity}
Our biggest source of ambiguity during template creation was singular \textit{they}, as \textit{they} is also a third person plural pronoun.
For example, if an \textit{advisor} and \textit{student} were meeting to discuss \textit{their} future, this could potentially refer to their future \textit{together}.
This problem applied across grammatical cases.
In addition, possessive sentences were potentially ambiguous across all pronoun series; when discussing a \textit{doctor} and a \textit{patient} and someone's diagnosis, this could be the \textit{doctor}'s diagnosis (i.e., the diagnosis made by the doctor), or the \textit{patient}'s diagnosis (i.e., the diagnosis the patient received).
All ambiguous templates were discarded and subsequently reworked.

\paragraph{Verbal Constraints}
The structural constraint of template pairs being identical until the pronoun led to some difficulties in finding appropriate (logically and semantically plausible) endings for the two sentences, particularly with accusative pronouns.
With nominative pronouns, we had to ensure we used verbs in the past tense and avoid was/were, so that our templates could be used with both \textit{he/she/xe} and singular \textit{they}.
It was also sometimes difficult to create single-entity sentences that were semantically close to the double-entity versions because the latter only made sense with two entities (e.g., ``X gave Y something'').\looseness=-1

\subsection{Data Validation}
\label{sec:data-validation}

As \textsc{WinoPron} templates have structural constraints that can be programmatically validated, we wrote automatic checks for these.
In addition, we performed human annotation of the sentences for grammaticality, and unique, correct coreferences.

\paragraph{Automatic Checks}
We automatically checked our data for completeness first, i.e., that every occupation-participant pair had sentence templates for nominative, accusative, and possessive pronouns.
We then automatically checked structural constraints, e.g., that a pair of templates must always be identical until the pronoun slot, and that no additional pronouns appeared in the sentence.

\paragraph{Human Annotation}

Both authors who created the schemas systematically annotated them, rating 100\% of the final instances as grammatical and 100\% of them as having unique, correct coreferences.
We confirmed the uniqueness of coreferences by marking each data instance as coreferring with the appropriate antecedent and \textit{not} coreferring with the other antecedent.
An additional annotator independently verified the final templates, rating 100\% of them as grammatical, and 98.2\% as having unique, correct coreferences.

\begin{figure*}[t!]
    \centering
    \includegraphics[width=\linewidth]{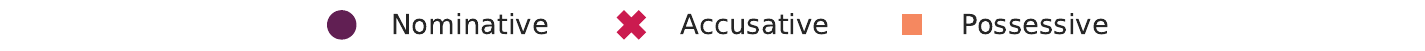}
    \includegraphics[width=0.8\linewidth]{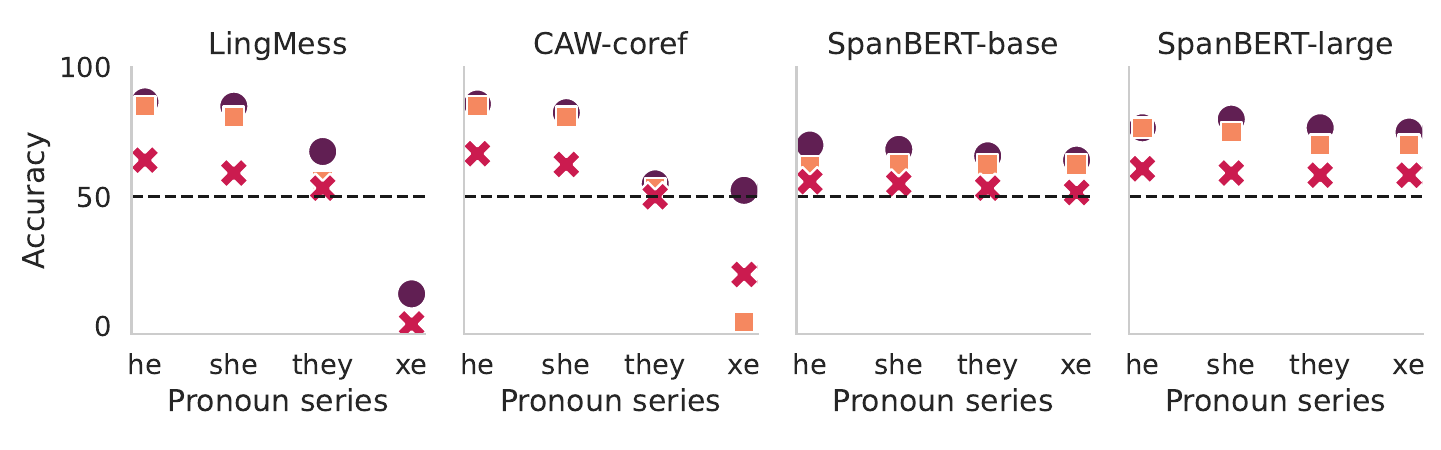}
    \includegraphics[width=\linewidth]{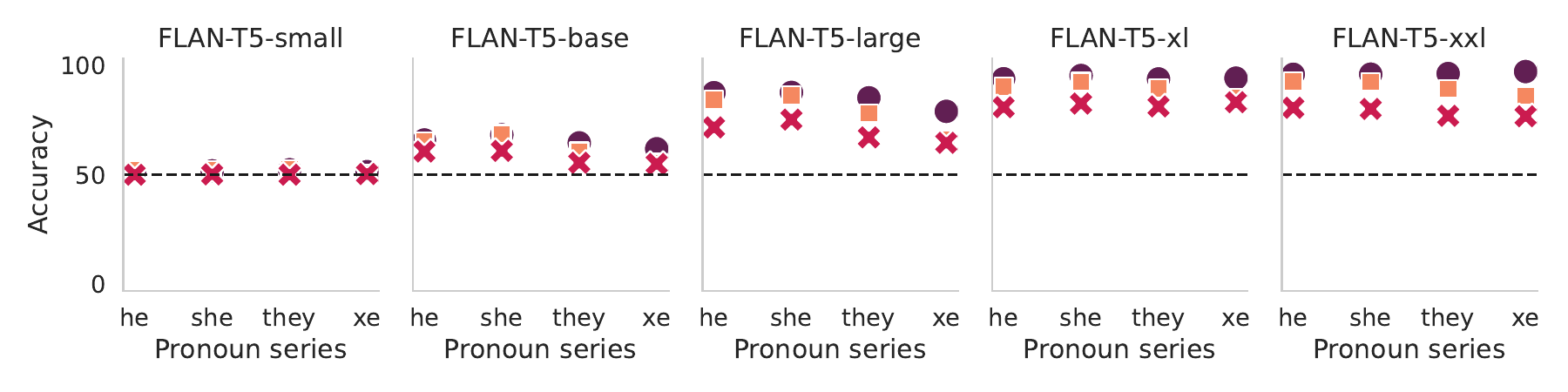}
    \caption{Accuracy on \textsc{WinoPron} by case and pronoun series with supervised coreference resolution systems (CAW-coref and LingMess), and language models fine-tuned for coreference resolution (SpanBERT) and prompted zero-shot (FLAN-T5), compared to random performance (50\%). Accusative pronoun performance is worse than other grammatical cases, and singular \textit{they} and the neopronoun \textit{xe} are challenging for several models.}
    \label{fig:results}
\end{figure*}

\section{Performance and Consistency}
\label{sec:performance-consistency}

To demonstrate the effects of our changes, we evaluate performance and consistency on \textsc{WinoPron} with a range of models with different levels of training for coreference resolution.

\subsection{Models}
\label{sec:models}

\noindent\textbf{LingMess}~\citep{otmazgin-etal-2023-lingmess} is a state-of-the-art, linguistically motivated, mixture-of-experts system for coreference resolution. \\
\vspace{-3mm}

\noindent\textbf{CAW-coref}~\citep{doosterlinck-etal-2023-caw} is a state-of-the-art word-level coreference resolution system based on an encoder-only model. \\
\vspace{-3mm}

\noindent\textbf{SpanBERT}~\citep{joshi-etal-2020-spanbert} is an encoder-only language model pre-trained with a span prediction objective and further enhanced for coreference resolution with fine-tuning data. We use both available model sizes (base and large) for evaluation. \\
\vspace{-3mm}

\noindent\textbf{FLAN-T5}~\citep{chung-flan-t5} is an instruction-tuned language model which is not trained for coreference resolution.
We evaluate on five model sizes (small, base, large, xl, and xxl), with prompts from the FLAN collection~\citep{flancollection23}. See Appendix \ref{sec:prompting} for details on prompting.

\subsection{Performance Results}
\label{sec:performance}

We first show how our changes affect overall performance between Winogender Schemas and \textsc{WinoPron}.
Then we use \textsc{WinoPron} to investigate differences across case (which we have balanced for) and pronoun sets (which can now be evenly compared).
Additional results are in Appendix \ref{sec:additional-results}.

\begin{table}[t]
    \centering
    \begin{tabularx}{\linewidth}{Xrrr}
        \toprule
        \textbf{System} & \textbf{WS} & \textbf{WP} & \textbf{$\Delta$F$_1$} \\
        \midrule
        LingMess & 85.5 & 64.4 & \textcolor{red}{-21.1} \\
        CAW-coref & 81.3 & 67.3 & \textcolor{red}{-14.0} \\
        \midrule
        SpanBERT-base & 71.8 & 61.6 & \textcolor{red}{-10.2} \\
        SpanBERT-large & 82.0 & 70.1 & \textcolor{red}{-11.9} \\
        \midrule
        FLAN-T5-small & 52.2 & 51.6 & \textcolor{red}{-0.6} \\
        FLAN-T5-base & 66.6 & 62.4 & \textcolor{red}{-4.2} \\
        FLAN-T5-large & 89.2 & 78.0 & \textcolor{red}{-11.2} \\
        FLAN-T5-xl & 97.4 & 89.0 & \textcolor{red}{-8.4} \\
        FLAN-T5-xxl & 97.5 & 88.8 & \textcolor{red}{-8.7} \\
        \bottomrule
    \end{tabularx}
    \caption{Overall performance (F$_1$) of coreference resolution systems on Winogender Schemas (WS) and \textsc{WinoPron} (WP). \textsc{WinoPron} is harder for all systems.}
    \label{tab:results-winogender-vs-winopron}
    \vspace{-2mm}
\end{table}

\paragraph{\textsc{WinoPron} is harder than Winogender Schemas.}
As Table \ref{tab:results-winogender-vs-winopron} shows, all the systems we evaluate perform worse on \textsc{WinoPron}, with F1 dropping on average by 10 percentage points compared to Winogender Schemas.
Patterns of performance across models are similar between Winogender Schemas and \textsc{WinoPron}, with similar scaling behaviour for both SpanBERT and FLAN-T5.
Notably, scale seems to supercede supervision, as the largest FLAN-T5 models perform the best overall.
Smaller FLAN-T5 models perform at chance level, which is likely a reflection of the ``demand gap'' induced through prompting~\citep{hu2024auxiliary}.

\paragraph{Accusative pronouns are harder.}
When model accuracy is split by grammatical case and pronoun series, we see that \textit{all} models struggle with accusative pronouns.
In general, systems perform best at resolving nominative pronouns, with a slight decrease for possessive pronouns and a large drop for accusative pronouns, as seen in Figure \ref{fig:results}.
This finding holds even for the best performing models on \textsc{WinoPron}, FLAN-T5-xl and FLAN-T5-xxl, where accuracy with accusative pronouns (81.9\% and 78.6\%) is much lower than with nominative (94.3\% and 96.3\%) or possessive (89.3\% and 90.0\%) pronouns.
We hypothesize that the performance gap for accusative pronouns is partially an effect of frequency; \textit{him} tokens appear roughly half as often in large pre-training corpora as \textit{he} and \textit{his} tokens~\citep{elazar2024whats}.

\paragraph{Performance with singular \textit{they} and neopronouns is bimodal.}
For the supervised coreference resolution systems (LingMess and CAW-coref), performance with singular \textit{they} is close to chance, and performance with the neopronoun \textit{xe} is far below chance, despite good performance with \textit{he/him/his} and \textit{she/her/her}.
SpanBERT performance also shows a gap between singular \textit{they} and neopronoun performance compared to data-rich pronouns, although the gap is much smaller.
These findings mirror those of \citet{cao-daume-iii-2020-toward,lauscher-etal-2022-welcome} and \citet{gautam2024robustpronounfidelityenglish}.
However, in contrast to \citeposs{gautam2024robustpronounfidelityenglish} findings with encoder-only and decoder-only models, there is no large difference in accuracy across pronoun sets with FLAN-T5 models.
As FLAN-T5 has been instruction fine-tuned for the task of coreference resolution but not pronoun fidelity~\citep{chung-flan-t5}, this could explain the model's ability to generalize to new pronouns in our setting.

\subsection{Consistency Results}
Next, we evaluate system consistency on groups of closely related instances in \textsc{WinoPron}, in order to dissect performance results and examine if systems are really right for the right reasons.
We follow \citet{ravichander-etal-2022-condaqa} in operationalizing consistency by taking the score of the lowest-performing instance in the group as the group's score.
We consider two groups, illustrated in Figure \ref{fig:consistency-grouping}: (a) \textit{pronoun consistency}, and (b) \textit{disambiguation consistency}, inspired by \citeposs{abdou-etal-2020-sensitivity} pair accuracy on Winograd Schemas.
In both cases, we report the percentage of groups for which a model performs consistently.\looseness=-1

Pronoun consistency measures model robustness across pronoun sets, i.e., if a model fails with even one pronoun set on a given template, then its score for that template is zero.
As we consider four pronoun sets, chance is $50\%^4$, or $6.25\%$.
Disambiguation consistency measures a system's ability to resolve a fixed pronoun to competing antecedents in paired templates.
Chance is thus $0.5^2$, or $0.25$.\looseness=-1

\begin{figure}[t]
    \begin{cframed}
    \centerline{\textbf{Pronoun consistency}}
    
    \noindent (a) \textbf{The counselor} informed the patient that \textit{his} qualifications were in psychology.

    \noindent (b) \textbf{The counselor} informed the patient that \textit{her} qualifications were in psychology.

    \noindent (c) \textbf{The counselor} informed the patient that \textit{their} qualifications were in psychology.

    \noindent (d) \textbf{The counselor} informed the patient that \textit{xyr} qualifications were in psychology.
    
    \vspace{0.05in}
    \Hrule[gray!40]{3pt}{5pt}
    \vspace{0.05in}

    \centerline{\textbf{Disambiguation consistency}}
    
    \noindent (a) \textbf{The counselor} informed the patient that xyr \textit{qualifications were in psychology}.
    
    \noindent (b) The counselor informed \textbf{the patient} that xyr \textit{insurance covered the cost of the sessions}.
    
    \end{cframed}
    \vspace{-2mm}
    \caption{Example groups for scoring consistency metrics using \textsc{WinoPron} templates for \textit{counselor}, \textit{patient} and possessive pronouns, with the antecedent bolded.}
    \label{fig:consistency-grouping}
\end{figure}

\begin{table}[th!]
    \centering
    \begin{tabularx}{\linewidth}{Xrr}
        \toprule
        \textbf{Model} & \textbf{PronounC} & \textbf{DisambigC} \\
        \midrule
         LingMess & \textcolor{red}{\textit{4.2}} & 33.3 \\
        CAW-coref & 18.3 & 34.7 \\
        \midrule
        SpanBERT-base & 50.0 & \textcolor{red}{\textit{24.3}} \\
        SpanBERT-large & \textbf{60.0} & 41.2 \\
        \midrule
        FLAN-T5-small & \textcolor{red}{\textit{3.9}} & \textcolor{red}{\textit{0.0}} \\
        FLAN-T5-base & \textcolor{red}{\textit{0.8}} & \textcolor{red}{\textit{0.0}} \\
        FLAN-T5-large & 14.4 & \textcolor{red}{\textit{5.4}} \\
        FLAN-T5-xl & 55.3 & \textbf{51.4} \\
        FLAN-T5-xxl & 43.9 & 43.3 \\
        \bottomrule
    \end{tabularx}
    \caption{Consistency results on \textsc{WinoPron}.
    Chance is 6.25\% for pronoun consistency (PronounC) and 25\% for disambiguation consistency (DisambigC).
    \textcolor{red}{\textit{Red, italicized numbers}} are worse than chance.}
    \label{tab:results-consistency}
\end{table}

\paragraph{SpanBERT-large is more robust to pronoun variation.}
As Table \ref{tab:results-consistency} shows, LingMess and the small and base sizes of FLAN-T5 score below chance, the former due to near-zero performance on \textit{xe/xem/xyr}, and the latter due to poor performance overall.
Interestingly, SpanBERT-large is more consistent (60.0\%) than FLAN-T5-xl (55.3\%) and FLAN-T5-xxl (43.9\%).
This indicates that despite its lower overall performance in Section \ref{sec:performance}, SpanBERT-large is more robust to pronominal variation.

\paragraph{The best model can only disambiguate half of the sentence pairs.}
Following from its high overall performance, FLAN-T5-xl has the highest disambiguation consistency score at 51.4\%, just over half the template pairs we evaluate.
In contrast, SpanBERT-base has disambiguation consistency below chance (24.3\%).
Given its reasonable overall performance, this result could stem from model bias, i.e., over-resolving a pronoun to a particular antecedent, disregarding the disambiguating context.
We thus investigate bias in more detail next.

\section{Pronominal Bias}
\label{sec:bias}

So far, we have focused on coreference resolution performance and consistency and found that 
accusative forms and less frequent pronoun sets are harder, and models are mostly non-robust to pronominal variation and antecedent disambiguation.
However, we have not established the extent to which models fail because they simply cannot perform the task, or if they are over-resolving a pronoun to a particular antecedent due to biased associations between them.
Thus, we aim to disentangle performance and bias in this section.

Winogender Schemas were originally proposed to measure gender bias in coreference resolution by using pronouns (a form of lexical gender) as a proxy for social gender.
\citet{rudinger-etal-2018-gender} then correlate incorrect resolution of English masculine and feminine pronouns with occupational statistics from the USA.
By conflating lexical and social gender (see \citet{cao-daume-iii-2021-toward} for a critical discussion), their analysis is subject to the same limitations as their data: treating different grammatical cases of the same pronoun as equivalent, and focusing only on \textit{he} and \textit{she}.
We thus propose a new method for evaluating pronominal bias in coreference resolution, correcting for these issues, and we then apply our method to investigate bias in SpanBERT models on \textsc{WinoPron}.\looseness=-1

\subsection{Evaluating Pronominal Bias}

When proposing a new method to evaluate pronominal bias in coreference resolution systems, our primary goal is to disentangle performance and bias.
In other words, we should have reason to believe that the model can perform the task, and that the reason it gets an instance wrong is specifically due to bias.
Additionally, we would like our method to work with an arbitrary set of pronouns of interest, and multiple surface forms of those pronouns.

\paragraph{Measuring Performance}
We first \textbf{(1)} isolate template pairs where the system attempts the task of coreference resolution as intended, i.e., the system resolves each pronoun to the occupation or participant (regardless of correctness).
Next, we \textbf{(2)} focus on the template pairs that the model can \textit{correctly} disambiguate with at least one pronoun set, $p_a$.
We deem the model capable of performing coreference resolution on this set of template pairs if it can resolve them with at least one pronoun set.

\paragraph{Measuring Bias}
Of the template pairs that a model can successfully disambiguate with at least one pronoun $p_a$, we then \textbf{(3)} focus on cases where the model fails to disambiguate the exact same template pair with a different pronoun $p_b \neq p_a$, as this is likely due to bias.
If the model over-resolves $p_b$ to the occupation, we posit that the model has a \textit{positive bias} between $p_b$ and that occupation.
On the other hand, if it over-resolves $p_b$ to the participant, the model is biased against associating $p_b$ with the occupation, i.e., it has a \textit{negative bias}.

\paragraph{Comparing Results}
With sets of positively and negatively biased occupations for each pronoun form, we want to quantify how many of a model's reasonable attempts to resolve a pronoun gave biased outputs.
We thus compute the percentage of templates that result in bias (see Measuring Bias) of the total templates that a model attempts to resolve with that pronoun, given  that it can correctly solve it with at least one pronoun (see Measuring Performance).
This gives us a quantitative measure of ``how biased'' a model is which also controls for whether a model is attempting the task and can perform the task with another pronoun.
In addition, we can quantify whether two models or two surface forms of a pronoun set have similar occupational biases by computing the Jaccard index~\citep{jaccard1912}, i.e., the size of the intersection of the biased occupation sets divided by the size of their union.

\begin{table*}[th]
    \centering
    \begin{tabularx}{\linewidth}{cGXGXGX}
        \toprule
        \multirow{2}*{\textbf{Pronouns}} & \multicolumn{2}{c}{\textbf{Nominative case}} & \multicolumn{2}{c}{\textbf{Accusative case}} & \multicolumn{2}{c}{\textbf{Possessive case}} \\
         & \multicolumn{1}{c}{\textbf{Positive}} & \multicolumn{1}{c}{\textbf{Negative}} & \multicolumn{1}{c}{\textbf{Positive}} & \multicolumn{1}{c}{\textbf{Negative}} & \multicolumn{1}{c}{\textbf{Positive}} & \multicolumn{1}{c}{\textbf{Negative}} \\
        \midrule
        \multirow{2}*{he/him/his} & \textit{engineer} \textit{painter} & \textit{receptionist} \textit{secretary} & \multirow{2}*{\centerline{--}} & \textit{dietitian} \textit{secretary} & \textit{practitioner} \textit{chef} & \textit{hairdresser}
              \textit{secretary} \\
        \midrule
        \multirow{2}*{she/her/her} & 
        \textit{hairdresser} \textit{painter} & \textit{accountant} \textit{plumber} & \multirow{2}*{\textit{cashier}} & \textit{firefighter} \textit{mechanic} & \textit{practitioner} \textit{painter} & \textit{accountant} \textit{surgeon} \\
        \midrule
        \multirow{2}*{they/them/their} & 
         \multirow{2}*{\centerline{--}} & \textit{accountant} \textit{plumber} & \multirow{2}*{\centerline{--}} & \textit{cashier} \phantom{b} \textit{dietitian} & \textit{advisor} \textit{baker} & \textit{accountant} \textit{surgeon} \\
        \midrule
        \multirow{2}*{xe/xem/xyr} & 
        \multirow{2}*{\centerline{--}} & \textit{hairdresser} \textit{engineer} & \multirow{2}*{\centerline{--}} & \textit{mechanic} \textit{cashier} &  \textit{advisor} \textit{baker} & \textit{engineer} \textit{supervisor} \\
        \bottomrule
    \end{tabularx}
    \caption{A sample of SpanBERT-large's biases when resolving pronouns to occupations. Positive bias: the model over-resolves the pronoun to that occupation. Negative bias: the model under-resolves the pronoun to the occupation.}
    \vspace{-2mm}
    \label{tab:bias}
\end{table*}

\subsection{Results}

\begin{figure}[t!]
    \centering
    \includegraphics[width=\linewidth]{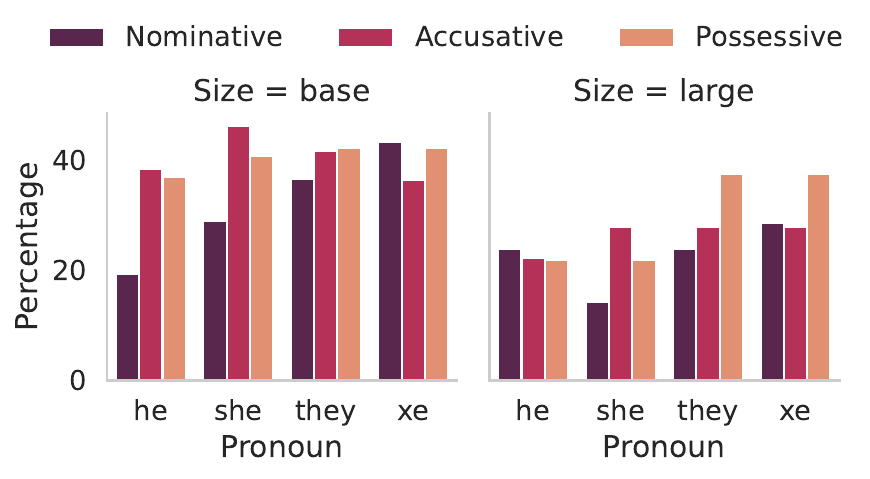}
    \caption{Percentage of model-attempted templates that show bias, for SpanBERT-base and SpanBERT-large.}
    \label{fig:bias}
\end{figure}

We apply our method to SpanBERT-base and SpanBERT-large and collect all instances of positive and negative bias between a pronoun form and an occupation.
Aggregated bias results for both models are shown in Figure \ref{fig:bias}, and Table \ref{tab:bias} shows a sample of biased occupations for SpanBERT-large.

\paragraph{SpanBERT-base is more biased than SpanBERT-large.}
As Figure \ref{fig:bias} shows, a larger percentage of SpanBERT-base's attempted and resolvable templates show biased behaviour when compared to SpanBERT-large.
This pattern holds even when examining positive and negative biases separately.
However, there are more negatively biased occupations than positively biased ones for both models.

\begin{table}[t!]
  \centering
  \begin{tabularx}{\linewidth}{Xrrrr}
    \toprule 
    \textbf{Grammatical case} &
    \textbf{he} & \textbf{she} & \textbf{they} & \textbf{xe}  \\
    \midrule
    Nominative & 0.14 & 0.15 & 0.17 & 0.32 \\
    Accusative & 0.12 & 0.10 & 0.25 & 0.29 \\
    Possessive & 0.12 & 0.18 & 0.24 & 0.24 \\
    \bottomrule
  \end{tabularx}
   \caption{Similarity of biased occupations between SpanBERT-base and SpanBERT-large, quantified with the Jaccard index (0.0 -1.0; higher is more similar).}
 \label{tab:bias-similarity-models}
 \vspace{-3mm}
\end{table}

\paragraph{Bias is qualitatively different across model sizes.}
In addition to being quantitatively different, we find that despite being trained and fine-tuned on the same data, there is low overlap between the occupational biases acquired by SpanBERT-base and SpanBERT-large (see Table \ref{tab:bias-similarity-models}).
For instance, the former positively associates \textit{she} with \textit{machinist}, while the latter positively associates \textit{she} with \textit{hairdresser} and \textit{painter}.
Only \textit{they/them/their} and \textit{xe/xem/xyr} have slightly higher overlap, mostly due to negative bias, as these models under-resolve these particular pronouns to all occupations.

\begin{table}[t!]
  \centering
  \begin{tabularx}{\linewidth}{Xrrrr}
    \toprule 
    \textbf{Case pairings} &
    \textbf{he} & \textbf{she} & \textbf{they} & \textbf{xe}  \\
    \midrule
    \rowcolor{grey} \multicolumn{5}{c}{SpanBERT-base} \\
    Nom-Acc & 0.10 & 0.00 & 0.00 & 0.00 \\
    Acc-Poss & 0.07 & 0.13 & 0.14 & 0.07 \\
    Nom-Poss & 0.07 & 0.11 & 0.10 & 0.09 \\
    \midrule
    \rowcolor{grey} \multicolumn{5}{c}{SpanBERT-large} \\
    Nom-Acc & 0.10 & 0.00 & 0.07 & 0.06 \\
    Acc-Poss & 0.22 & 0.00 & 0.06 & 0.06 \\
    Nom-Poss & 0.17 & 0.29 & 0.15 & 0.19 \\
    \bottomrule
  \end{tabularx}
   \caption{Similarity of biased occupations across pairings of grammatical case (nom: nominative, acc: accusative, poss: possessive) of a pronoun set, quantified with the Jaccard index (0.0 -1.0; higher is more similar).}
 \label{tab:bias-similarity-case}
 \vspace{-4mm}
\end{table}

\paragraph{Bias does not match qualitatively across grammatical case.}
In other words, positive bias with \textit{she} for an occupation does not entail positive bias with \textit{her}.
We quantify this systematically by computing Jaccard indices in Table \ref{tab:bias-similarity-case}, where we find that most pairings of grammatical case have very low overlap in their biases.
In fact, even contradictory associations are possible; SpanBERT-base has a positive bias between \textit{manager} and \textit{them}, but a negative bias betweeen \textit{manager} and \textit{their}.
Only nominative and possessive occupational biases in SpanBERT-large appear to somewhat consistently overlap with each other.
Although some of these instances (e.g., negative bias for \textit{secretary} with \textit{he}, \textit{him}, and \textit{his}) align with social stereotypes~\citep{haines-et-al-2016}, the overall pattern provides evidence that grammatical case in pronouns has its own set of biases that should be examined in their own right.~\looseness=-1

\paragraph{Bias is not additive.}
Even though SpanBERT-large has positive bias for \textit{baker} and \textit{her}, \textit{their} as well as \textit{xyr}, this does not imply that the model must have a negative bias between \textit{baker} and \textit{his}; it does not.
This further highlights the need for evaluation that goes beyond binary, oppositional operationalizations of gender via pronouns.

\section{Discussion}

By systematically identifying and fixing issues with Winogender Schemas~\citep{rudinger-etal-2018-gender}, we create a new dataset, \textsc{WinoPron}, and find that: \textbf{(1)} different grammatical cases of pronouns show vastly different performance and bias characteristics, \textbf{(2)} pronominal biases are rich and varied, of which \textit{he} and \textit{she} are only the tip of the iceberg, and \textbf{(3)} model biases are complex and do not necessarily match our intuitions about them. Based on our findings, we make some recommendations for researchers who study coreference resolution and those who study bias and fairness via pronouns.

First, grammatical case is a dimension of pronominal performance and bias that warrants more study~\citep{munro-morrison-2020-detecting}.
In particular, we hope that future work further investigates \textit{why} accusative pronouns are harder.
The patterns we demonstrate (both for performance and bias) could arise from a number of sources beyond mere frequency, including quirks of our dataset, or the distribution of semantic roles in training data for coreference resolution systems.

Second, we echo prior calls for fairness researchers to attend to the differences between social gender and terms that index it~\citep{cao-daume-iii-2021-toward,gautam-etal-2024-stop}, to include more diversity in pronouns~\citep{baumler-rudinger-2022-recognition,lauscher-etal-2022-welcome,hossain-etal-2023-misgendered}, and to move towards richer operationalizations of gender~\citep{devinney-et-al-2022,ovalle2023m} and bias~\citep{blodgett-etal-2020-language}.
Specifically, future work on bias in coreference resolution should treat pronominal bias as distinct from (social) gender bias, defend how and why pronouns are mapped to social gender, and move beyond binary, oppositional methods of evaluation.

Lastly, as our work is a case study in how careful data curation and operationalization affects claims about system performance and bias, we emphasize the need for thoughtful data work~\citep{sambasivan-et-al-2021}, and encourage the use of automatic checks when feasible, as in our work.

\section{Related work}

Besides \citet{rudinger-etal-2018-gender}, there are a number of papers that tackle gender bias in coreference resolution, all of which differ from ours.
Similar to Winogender Schemas, WinoBias~\citep{zhao-etal-2018-gender} proposes Winograd-like schemas that focus on occupations to evaluate gender bias in coreference resolution.
However, WinoBias only covers \textit{he} and \textit{she}, rather than our coverage of all English pronoun sets by design.
In addition, like Winogender, WinoBias also treats pronouns in all grammatical cases the same way.
WinoNB schemas~\citep{baumler-rudinger-2022-recognition} evaluate how coreference resolution systems handle singular they and plural they with similar schemas.
Beyond these constructed schemas, there also exist datasets of challenging sentences found ``in the wild,'' such as BUG~\citep{levy-etal-2021-collecting-large}, GAP~\citep{webster-et-al-2018}, and GICOREF~\citep{cao-daume-iii-2021-toward}.
However, as these natural datasets are not carefully constructed like Winograd-like schemas, pronouns cannot be swapped in dataset instances and still be assumed to be grammatical or coherent.

Our work is also one among several papers that investigate datasets for problems including low quality or noisy data~\citep{elazar2024whats,abela-etal-2024-tokenisation}, artifacts~\citep{shwartz-etal-2020-grounded,herlihy-rudinger-2021-mednli,elazar-etal-2021-back,dutta-chowdhury-etal-2022-towards}, contamination~\cite{balloccu-etal-2024-leak,deng-etal-2024-investigating}, and issues with conceptualization and operationalization of bias~\citep{blodgett-etal-2021-stereotyping,selvam-etal-2023-tail,nighojkar-etal-2023-strong,subramonian-etal-2023-takes,gautam-etal-2024-stop}.
We cover many of these areas, but do not control for dataset artifacts, which we explain further in our \hyperref[sec:limitations]{Limitations} section.

\section{Conclusion}

We demonstrate a number of issues with the well-known Winogender Schemas dataset, which we fix in our new, expanded \textsc{WinoPron} dataset.
In addition, we propose a novel way to evaluate pronominal bias in coreference resolution that goes beyond the binary and focuses on lexical gender.
With our new dataset, we evaluate both supervised coreference resolution systems and language models, and find that the grammatical case of pronouns affects model performance and bias, and that bias varies widely across models, pronoun sets and grammatical cases.
Our work demonstrates that measurements of bias and robustness are only as good as the datasets and metrics we use to measure them, and we call for careful attention when developing future resources for evaluating bias and coreference resolution, with attention to grammatical case, more careful operationalizations of bias, and greater diversity in the pronouns we consider.

\section*{Limitations}
\label{sec:limitations}

As in Winogender Schemas, our schemas are not ``Google-proof'' and could conceivably be solved with heuristics, including word co-occurrences, which is a primary concern when creating and evaluating \textit{Winograd} schemas~\citep{levesque_et_all_2012,amsili-seminck-2017-google,elazar-etal-2021-back}.
The fact that we do not control for this means that our dataset gives \textit{generous} estimates of system performance, particularly for strong language models like FLAN-T5, but it also means that this dataset is inappropriate to test ``reasoning.''
Our dataset construction instead controls for simple system heuristics that are relevant for coreference resolution, such as always picking the first entity in the sentence, or always picking the second.

We take steps to prevent data contamination~\citep{jacovi-etal-2023-stop}, including not releasing our data in plain text, and not evaluating with language models behind closed APIs that do not guarantee that our data will not be used to train future models~\citep{
balloccu-etal-2024-leak}. However, as we cannot guarantee a complete absence of data leakage unless we never release the dataset, we encourage caution in interpreting results on \textsc{WinoPron} with models trained on data after August 2024.

Finally, we note that as our evaluation set only contains one set of templates per occupation-participant pair, our results represent a point in the distribution of bias related to that occupation.
We thus echo \citeauthor{rudinger-etal-2018-gender}'s \citeyearpar{rudinger-etal-2018-gender} view of Winogender Schemas as having ``high positive predictive value and low negative predictive value'' for bias.
In other words, they may demonstrate evidence of pronominal bias in systems, but not prove its absence.
In the case of large language models in particular, using a small number of templates for templatic evaluation is known to be brittle even to small, meaning-preserving changes to the template~\citep{quantifyingsocialbiases2022,selvam-etal-2023-tail}.
Our dataset's small size is a result of us requiring a tightly controlled and structured dataset to evaluate how coreference resolution varies.
Thus, it may differ from realistic examples (which would have other differences that confound bias results).
We wish to emphasize that in addition to controlled datasets like ours, realistic evaluation is also necessary for holistically evaluating performance, robustness and bias in coreference resolution.

\section*{Acknowledgements}
The authors thank Timm Dill for several rounds of patient annotation, and are grateful to Rachel Rudinger, Benjamin Van Durme, and our CRAC reviewers for their comments.
Vagrant Gautam received funding from the BMBF’s (German Federal Ministry of Education and Research) SLIK project under the grant 01IS22015C.
Anne Lauscher's work is funded under the Excellence Strategy of the German Federal Government and States.

\bibliography{custom}
\bibliographystyle{acl_natbib}

\appendix

\section{List of Occupations}
\label{sec:occupation-list}

The occupations along with their respective participants in parentheses are listed below in alphabetical order. This list is identical to the occupations and participants in \citet{rudinger-etal-2018-gender}, except that we pair examiner with intern rather than victim: \\

\noindent accountant (taxpayer), administrator (undergraduate), advisor (advisee), appraiser (buyer), architect (student), auditor (taxpayer), baker (customer), bartender (customer), broker (client), carpenter (onlooker), cashier (customer), chef (guest), chemist (visitor), clerk (customer), counselor (patient), dietitian (client), dispatcher (bystander), doctor (patient), educator (student), electrician (homeowner), engineer (client), examiner (intern), firefighter (child), hairdresser (client), hygienist (patient), inspector (homeowner), instructor (student), investigator (witness), janitor (child), lawyer (witness), librarian (child), machinist (child), manager (customer), mechanic (customer) nurse (patient), nutritionist (patient), officer (protester), painter (customer), paralegal (client), paramedic (passenger), pathologist (victim), pharmacist (patient), physician (patient), planner (resident), plumber (homeowner), practitioner (patient), programmer (student), psychologist (patient), receptionist (visitor), salesperson (customer), scientist (undergraduate), secretary (visitor), specialist (patient), supervisor (employee), surgeon (child), teacher (student), technician (customer), therapist (teenager), veterinarian (owner), worker (pedestrian) 

\section{Annotator Demographics}
\label{sec:annotator-information}

All three annotators (two authors and an additional annotator) are fluent English speakers.
The two authors who create and validate templates have linguistic training at the undergraduate level.
One author and one annotator have experience with using singular \textit{they} and neopronouns, while the other author has prior exposure to singular \textit{they} but not the neopronoun \textit{xe}.

\section{Annotation Instructions}
\label{sec:annotation-instructions}

\subsection{Task 1 Description}

Together with this annotation protocol, you have received a link to a Google Sheet. The sheet contains 2 data columns and 2 task columns of randomized data. The data columns consist of
\begin{itemize}
    \item Sentences which you are asked to annotate for grammaticality; and
    \item Questions about pronouns in the sentence, which you are asked to answer
\end{itemize}
Please be precise in your assignments and do not reorder the data. The columns have built-in data validation and we will perform further tests to check for consistent annotation.

\subsubsection{Grammaticality}
In the “Grammatical?” column, please enter your grammaticality judgments of the sentence, according to Standard English. The annotation options are:
\begin{itemize}
    \item \textbf{grammatical} (for fluent, syntactically valid and semantically plausible sentences)
    \item \textbf{ungrammatical} (for sentences that have any typos, grammatical issues, or if the sentence describes a situation that don’t make sense, or just sounds weird)
    \item \textbf{not sure} (if you are not sure whether it is clearly grammatical or ungrammatical)
\end{itemize}
Examples:
\begin{itemize}
    \item \textit{The driver told the passenger that he could pay for the ride with cash.}\\
    => grammatical
    \item \textit{The driver said the passenger that he could pay for the ride with cash.}\\
    => ungrammatical (because ‘said’ is intransitive in Standard English)
\end{itemize}

\subsubsection{Questions about pronouns}
Every sentence contains a pronoun, and the “Question” column asks whether it refers to a person mentioned in the sentence or not. The annotation options are:
\begin{itemize}
    \item \textbf{yes} (if the pronoun refers to the person)
    \item \textbf{no} (if the pronoun does not refer to the person)
    \item \textbf{not sure} (if you are not sure about whether the pronoun refers to the person)
\end{itemize}
Examples: 
\begin{itemize}
    \item \textit{The driver told the passenger that he could pay for the ride with cash.}\\
    Does the pronoun he refer to the driver?\\
    => no
    \item \textit{The driver told the passenger that he could pay for the ride with cash.}\\
    Does the pronoun he refer to the passenger?\\
    => yes
\end{itemize}

\subsection{Task 2 Description}
Together with this annotation protocol, you have received a link to a Google Sheet. The sheet contains 1 randomized data column and 1 task column.\\
Each row in the data column consists of multiple sentences, of which precisely one sentence contains a blank. Your task is to determine the appropriate pronoun to fill in the blank, and enter it in the “Pronoun” column. Here, appropriate means correct in both form and case.\\
The tasks are designed to be unambiguous, so please provide only one solution and do not reorder the data.

Example: 
\begin{itemize}
    \item \textit{The driver felt unhappy because he did not make enough money. The driver wondered whether} {\_}{\_}{\_} \textit{should take out a loan.}\\
    => he
\end{itemize}

\section{Prompting}
\label{sec:prompting}

\begin{table*}[ht!]
\centering
    \begin{tabularx}{\linewidth}{lX}
    \toprule
    \textbf{ID}  & \textbf{Template}  \\ 
    \midrule
    0      
             &  \texttt{\{task\}}\slashn\slashn\texttt{\{options\}}\slashn Who is \texttt{\{pronoun\}} referring to? \\
    \midrule
    1      
             &  \texttt{\{task\}}\slashn\slashn Who is ``\texttt{\{pronoun\}}'' in this prior sentence (see options)?\slashn\texttt{\{options\}} \\
    \midrule
    2      
             &  \texttt{\{task\}}\slashn\slashn Who is \texttt{\{pronoun\}} referring to in this sentence?\slashn\texttt{\{options\}} \\
    \midrule
    3      
             &  Choose your answer: \texttt{\{task\}}\slashn Tell me who \texttt{\{pronoun\}} is.\slashn\texttt{\{options\}} \\
    \midrule
    4      
             &  \texttt{\{task\}}\slashn Based on this sentence, who is \texttt{\{pronoun\}}?\slashn\slashn\texttt{\{options\}} \\
    \midrule
    5      
             &  Choose your answer: Who is \texttt{\{pronoun\}} in the following sentence?\slashn\slashn\texttt{\{task\}} \slashn\slashn\texttt{\{options\}} \\
    \midrule
    6      
             & Multi-choice problem: Which entity is \texttt{\{pronoun\}} this sentence?\slashn\slashn\texttt{\{task\}}  \slashn\slashn\texttt{\{options\}}  \\
    \midrule
    7      
             & Who is \texttt{\{pronoun\}} referring to in the following sentence?\slashn\texttt{\{task\}} \slashn\slashn\texttt{\{options\}}  \\
    \midrule
    8      
             & Note that this question lists possible answers. Which person is \texttt{\{pronoun\}} referring to in the following sentence?\slashn\texttt{\{task\}} \slashn\slashn\texttt{\{options\}}  \\
    \midrule
    9      
             &  \texttt{\{task\}}\slashn Who is ``\texttt{\{pronoun\}}''\slashn\texttt{\{options\}} \\
    \bottomrule
    \end{tabularx}
\caption{Prompting templates, where ``task'' is filled with each dataset instance, ``pronoun'' is the unique third person singular pronoun in that dataset instance, and ``options'' are the occupation and the participant.}
\label{tab:prompting-templates}
\end{table*}

Table~\ref{tab:prompting-templates} shows all 10 prompt templates we use to present our task instances to FLAN-T5.
Each template is presented in three variants to the model, where the options are changed:
\begin{enumerate}
    \item No options
    \item The occupation is presented first and the participant second
    \item The participant is presented first and the occupation second
\end{enumerate}

\section{Additional Results}
\label{sec:additional-results}

We report additional results on double- and single-entity sentences in \textsc{WinoPron}: $F_1$ scores in Table \ref{tab:results-f1}, precision in Table \ref{tab:results-precision}, and recall in Table \ref{tab:results-recall}.
Note that FLAN-T5 models generally perform worse on single-entity sentences compared to double-entity sentences because some of our prompts include options (see Section \ref{sec:prompting} for details) that confuse the model in this setting, despite being necessary to resolve double-entity sentences.

\begin{table*}[ht!]
    \centering
    \begin{tabularx}{\linewidth}{Xccccccccc}
        \toprule
        \multirow{2}*{\textbf{Data}} & \multirow{2}*{\textbf{LingMess}} & \multirow{2}*{\textbf{CAW-coref}} & \multicolumn{2}{c}{\textbf{SpanBERT}} & \multicolumn{5}{c}{\textbf{FLAN-T5}} \\
        & & & \textbf{base} & \textbf{large} & \textbf{small} & \textbf{base} & \textbf{large} & \textbf{xl} & \textbf{xxl} \\
        \midrule
        \rowcolor{grey} \multicolumn{10}{c}{Double-entity sentences} \\
        All         & 64.4 & 67.3 & 61.6 & 70.1 & 51.6 & 62.4 & 78.0 & \textbf{89.0} & 88.8 \\
        \midrule
        Nominative  & 73.5 & 77.6 & 67.2 & 77.2 & 51.9 & 65.4 & 85.1 & 94.7 & \textbf{96.7} \\
        Accusative  & 52.2 & 57.5 & 54.6 & 59.5 & 50.4 & 58.4 & 69.9 & \textbf{82.5} & 79.1 \\
        Possessive  & 67.4 & 66.5 & 62.9 & 73.6 & 52.3 & 63.4 & 79.1 & 89.7 & \textbf{90.7} \\
        \midrule
        \textit{he/him/his}      & 79.2 & 79.6 & 62.8 & 71.5 & 51.5 & 64.1 & 81.5 & 88.8 & \textbf{90.2} \\
        \textit{she/her/her}     & 76.3 & 76.6 & 62.1 & 71.6 & 51.5 & 66.1 & 83.3 & \textbf{90.6} & 89.9 \\
        \textit{they/them/their}    & 67.5 & 63.7 & 61.2 & 68.9 & 51.8 & 60.5 & 77.0 & \textbf{88.6} & 88.0 \\
        \textit{xe/xem/xyr}      & \textit{\textcolor{red}{8.5}} & \textit{\textcolor{red}{38.6}} & 60.4 & 68.5 & 51.4 & 58.7 & 70.3 & \textbf{88.0} & 87.3 \\
        \midrule
        \rowcolor{grey} \multicolumn{10}{c}{Single-entity sentences} \\
        All         & 73.2 & 75.6 & \textbf{95.5} & 88.0 & 77.3 & 76.3 & 81.5 & 83.1 & 84.3 \\
        \midrule
        Nominative  & 80.0 & 82.5 & \textbf{99.5} & 99.3 & 78.3 & 80.8 & 89.8 & 93.3 & 97.0 \\
        Accusative  & 61.1 & 65.0 & \textbf{87.3} & 67.5 & 76.2 & 69.6 & 69.8 & 70.1 & 66.5 \\
        Possessive  & 77.1 & 78.0 & \textbf{99.8} & 97.1 & 77.5 & 78.5 & 84.7 & 85.7 & 89.2 \\
        \midrule
        \textit{he/him/his}      & 92.7 & 94.3 & \textbf{94.7} & 85.6 & 77.6 & 81.3 & 86.8 & 88.2 & 88.6 \\
        \textit{she/her/her}     & 90.9 & 91.6 & \textbf{96.2} & 88.9 & 77.4 & 81.1 & 87.6 & 88.8 & 87.1 \\
        \textit{they/them/their}    & 75.2 & 69.8 & \textbf{96.0} & 88.7 & 79.3 & 76.1 & 84.3 & 85.7 & 86.8 \\
        \textit{xe/xem/xyr}      & 2.2 & 27.3 & \textbf{95.2} & 88.7 & 75.0 & 66.3 & 67.0 & 69.4 & 74.6 \\
        \bottomrule
    \end{tabularx}
    \caption{F$_1$ of coreference resolution systems on double- and single-entity sentences in \textsc{WinoPron}. We report F$_1$ overall, and split by grammatical case and pronoun set. \textcolor{red}{\textit{Red, italicized numbers}} are worse than chance (50.0 for double-entity sentences and not applicable for single-entity sentences).}
    \label{tab:results-f1}
\end{table*}

\begin{table*}[ht!]
    \centering
    \begin{tabularx}{\linewidth}{Xccccccccc}
        \toprule
        \multirow{2}*{\textbf{Data}} & \multirow{2}*{\textbf{LingMess}} & \multirow{2}*{\textbf{CAW-coref}} & \multicolumn{2}{c}{\textbf{SpanBERT}} & \multicolumn{5}{c}{\textbf{FLAN-T5}} \\
        & & & \textbf{base} & \textbf{large} & \textbf{small} & \textbf{base} & \textbf{large} & \textbf{xl} & \textbf{xxl} \\
        \midrule
        \rowcolor{grey} \multicolumn{10}{c}{Double-entity sentences} \\
        All         & 79.1 & 80.1 & 62.1 & 70.6 & 51.9 & 62.9 & 78.4 & \textbf{89.5} & 89.4 \\
        \midrule
        Nominative  & 88.3 & 88.7 & 67.4 & 77.4 & 52.1 & 65.7 & 85.4 & 95.1 & \textbf{97.1} \\
        Accusative  & 63.4 & 67.9 & 55.3 & 59.9 & 50.7 & 58.8 & 70.2 & \textbf{83.2} & 79.5 \\
        Possessive  & 86.1 & 83.6 & 63.5 & 74.3 & 52.8 & 64.3 & 79.6 & 90.2 & \textbf{91.5} \\
        \midrule
        \textit{he/him/his}      & 79.7 & 80.1 & 63.0 & 71.6 & 51.7 & 64.3 & 81.8 & 89.3 & \textbf{90.6} \\
        \textit{she/her/her}     & 77.6 & 77.9 & 62.3 & 71.8 & 51.7 & 66.3 & 83.6 & \textbf{91.1} & 90.3 \\
        \textit{they/them/their}    & 79.1 & 80.2 & 61.8 & 69.5 & 52.0 & 60.8 & 77.3 & \textbf{89.0} & 88.5 \\
        \textit{xe/xem/xyr}      & \textbf{100.0} & 88.1 & 61.3 & 69.3 & 52.1 & 60.1 & 70.7 & 88.6 & 88.0 \\
        \midrule
        \rowcolor{grey} \multicolumn{10}{c}{Single-entity sentences} \\
        All         & \textbf{100.0} & \textbf{100.0} & 96.0 & 88.4 & 78.9 & 77.6 & 82.4 & 84.0 & 85.6 \\
        \midrule
        Nominative  & \textbf{100.0} & \textbf{100.0} & \textbf{100.0} & \textbf{100.0} & 79.3 & 81.6 & 90.4 & 93.9 & 97.4 \\
        Accusative  & \textbf{100.0} & \textbf{100.0} & 88.1 & 67.9 & 77.5 & 70.5 & 70.7 & 71.1 & 68.1 \\
        Possessive  & \textbf{100.0} & \textbf{100.0} & 99.8 & 97.1 & 79.8 & 80.7 & 85.9 & 86.8 & 90.8 \\
        \midrule
        \textit{he/him/his}      & \textbf{100.0} & \textbf{100.0} & 95.0 & 86.0 & 78.6 & 81.9 & 87.5 & 88.9 & 89.5 \\
        \textit{she/her/her}     & \textbf{100.0} & \textbf{100.0} & 96.4 & 89.1 & 78.5 & 81.7 & 88.1 & 89.4 & 87.9 \\
        \textit{they/them/their}    & \textbf{100.0} & \textbf{100.0} & 96.4 & 89.1 & 80.3 & 76.9 & 85.2 & 86.5 & 87.9 \\
        \textit{xe/xem/xyr}      & \textbf{100.0} & \textbf{100.0} & 96.3 & 89.3 & 77.9 & 69.2 & 68.3 & 70.9 & 76.7 \\
        \bottomrule
    \end{tabularx}
    \caption{Precision on double- and single-entity sentences overall, and split by grammatical case and pronoun set. \textcolor{red}{\textit{Red, italicized numbers}} are worse than chance (50.0 for double-entity sentences, N/A for single-entity sentences).}
    \label{tab:results-precision}
\end{table*}

\begin{table*}[ht!]
    \centering
    \begin{tabularx}{\linewidth}{Xccccccccc}
        \toprule
        \multirow{2}*{\textbf{Data}} & \multirow{2}*{\textbf{LingMess}} & \multirow{2}*{\textbf{CAW-coref}} & \multicolumn{2}{c}{\textbf{SpanBERT}} & \multicolumn{5}{c}{\textbf{FLAN-T5}} \\
        & & & \textbf{base} & \textbf{large} & \textbf{small} & \textbf{base} & \textbf{large} & \textbf{xl} & \textbf{xxl} \\
        \midrule
        \rowcolor{grey} \multicolumn{10}{c}{Double-entity sentences} \\
        All         & 54.2 & 58.0 & 61.1 & 69.7 & 51.3 & 61.9 & 77.7 & \textbf{88.5} & 88.3 \\
        \midrule
        Nominative  & 62.9 & 69.0 & 67.1 & 77.1 & 51.8 & 65.2 & 84.8 & 94.3 & \textbf{96.3} \\
        Accusative  & \textit{\textcolor{red}{44.4}} & \textit{\textcolor{red}{49.8}} & 54.0 & 59.2 & 50.2 & 58.0 & 69.6 & \textbf{81.9} & 78.6 \\
        Possessive  & 55.4 & 55.2 & 62.3 & 72.9 & 51.9 & 62.5 & 78.7 & 89.3 & \textbf{90.0} \\
        \midrule
        \textit{he/him/his}      & 78.6 & 79.2 & 62.5 & 71.4 & 51.4 & 63.9 & 81.1 & 88.3 & \textbf{89.7} \\
        \textit{she/her/her}     & 75.0 & 75.3 & 61.9 & 71.4 & 51.4 & 65.9 & 83.0 & \textbf{90.1} & 89.5 \\
        \textit{they/them/their}    & 58.9 & 52.8 & 60.6 & 68.3 & 51.6 & 60.2 & 76.8 & \textbf{88.1} & 87.5 \\
        \textit{xe/xem/xyr}      & \textit{\textcolor{red}{4.4}} & \textit{\textcolor{red}{24.7}} & 59.4 & 67.8 & 50.8 & 57.4 & 69.9 & \textbf{87.5} & 86.6 \\
        \midrule
        \rowcolor{grey} \multicolumn{10}{c}{Single-entity sentences} \\
        All         & 57.8 & 60.8 & \textbf{95.1} & 87.6 & 75.9 & 75.0 & 80.6 & 82.1 & 83.1 \\
        \midrule
        Nominative  & 66.7 & 70.2 & \textbf{99.0} & 98.5 & 77.3 & 80.0 & 89.2 & 92.7 & 96.6 \\
        Accusative  & 44.0 & 48.1 & \textbf{86.5} & 67.1 & 74.9 & 68.7 & 69.0 & 69.1 & 65.0 \\
        Possessive  & 62.7 & 64.0 & \textbf{99.8} & 97.1 & 75.4 & 76.4 & 83.5 & 84.6 & 87.6 \\
        \midrule
        \textit{he/him/his}      & 86.4 & 89.2 & \textbf{94.4} & 85.3 & 76.5 & 80.8 & 86.1 & 87.4 & 87.8 \\
        \textit{she/her/her}     & 83.3 & 84.4 & \textbf{96.1} & 88.6 & 76.2 & 80.5 & 87.1 & 88.2 & 86.2 \\
        \textit{they/them/their}    & 60.3 & 53.6 & \textbf{95.6} & 88.3 & 78.3 & 75.3 & 83.5 & 85.0 & 85.8 \\
        \textit{xe/xem/xyr}      & 1.1 & 15.8 & \textbf{94.2} & 88.1 & 72.4 & 63.7 & 65.7 & 68.0 & 72.5 \\
        \bottomrule
    \end{tabularx}
    \caption{Recall on double- and single-entity sentences overall, and split by grammatical case and pronoun set. \textcolor{red}{\textit{Red, italicized numbers}} are worse than chance (50.0 for double-entity sentences, N/A for single-entity sentences)}
    \label{tab:results-recall}
\end{table*}

\end{document}